\documentclass[runningheads]{llncs}
\usepackage[T1]{fontenc}

\usepackage{graphicx}
\usepackage{csquotes}
\usepackage[acronym]{glossaries}
\usepackage{amsmath}
\usepackage{amssymb}

\newacronym{ai}{AI}{Artificial Intelligence}
\newacronym{car}{CAR}{Concept Activation Region}
\newacronym{cav}{CAV}{Concept Activation Vector}
\newacronym{cavlrp}{glCA}{global-to-local Concept Attribution}
\newacronym{crp}{CRP}{Concept-wise Relevance Propagation}
\newacronym{dnn}{DNN}{Deep Neural Network}
\newacronym{gbp}{GBP}{Guided Backpropagation}
\newacronym{gradcam}{GradCAM}{}
\newacronym{fcnn}{FasterRCNN}{}
\newacronym{lcrp}{L-CRP}{CRP for Localization Models}
\newacronym{lrp}{LRP}{Layer-wise Relevance Propagation}
\newacronym{net2vec}{Net2Vec}{}
\newacronym{nms}{NMS}{Non-Maximum Suppression}
\newacronym{nn}{NN}{neural network}
\newacronym{od}{OD}{object detection}
\newacronym{patcav}{PatCAV}{Pattern/Signal-CAV}
\newacronym{spatcav}{sPatCAV}{}
\newacronym{ssd}{SSD}{Single Shot MultiBox Detector}
\newacronym{tcav}{TCAV}{Testing with Concept Activation Vectors}
\newacronym{vae}{VAE}{variational autoencoder}
\newacronym{xai}{xAI}{eXplainable Artificial Intelligence}

\begin{document}
%
\title{Locally Testing Model Detections for Semantic Global Concepts}
%
%
\author{Franz Motzkus\thanks{Corresponding Author}\inst{1,2,3}\orcidID{0009-0009-4362-7907} \and\\
Georgii Mikriukov\inst{1,2}\orcidID{0000-0002-2494-6285} \and\\
Christian Hellert\inst{1}\orcidID{0000-0002-5781-6575} \and
Ute Schmid\inst{3}\orcidID{0000-0002-1301-0326}}
\authorrunning{F. Motzkus et al.}
%
\institute{Continental Automotive Technologies GmbH,\and Continental AI Lab Berlin\\
\email{franz.walter.motzkus@continental-corporation.com}
\email{\{firstname.lastname\}@continental-corporation.com}\and
University of Bamberg, Germany
\email{\{firstname.lastname\}@uni-bamberg.de}}
\maketitle              
\begin{abstract}

Ensuring the quality of black-box \glspl{dnn} has become ever more significant, especially in safety-critical domains such as automated driving.
While global concept encodings generally enable a user to test a model for a specific concept, linking global concept encodings to the local processing of single network inputs reveals their strengths and limitations.
Our proposed framework \gls{cavlrp} uses approaches from local (why a specific prediction originates) and global (how a model works generally) \gls{xai} to test \glspl{dnn} for a predefined semantical concept locally.
The approach allows for conditioning local, post-hoc explanations on predefined semantic concepts encoded as linear directions in the model's latent space.
Pixel-exact scoring concerning the global concept usage assists the tester in further understanding the model processing of single data points for the selected concept.
Our approach has the advantage of fully covering the model-internal encoding of the semantic concept and allowing the localization of relevant concept-related information.
The results show major differences in the local perception and usage of individual global concept encodings and demand for further investigations regarding obtaining thorough semantic concept encodings.


\end{abstract}

\section{Introduction}

While \glspl{dnn} have originally been introduced as black-box models due to their complexity and non-linearities, the rapidly evolving domain \gls{xai} seeks to make \gls{dnn} decision-making processes more transparent and comprehensible~\cite{samek_evaluating_2017,holzinger_explainable_2022,schwalbe_comprehensive_2023}.
The utility of \gls{xai} has been shown in various advancements ranging from simply explaining a single model output~\cite{holzinger_explainable_2022} to finding erroneous model behaviour~\cite{anders_finding_2022} and even improving and correcting the model itself~\cite{weber_beyond_2023}.
Especially for safety-critical applications in, e.g., autonomous mobility~\cite{schwalbe_structuring_2020} or the medical domain~\cite{combi_manifesto_2022} shortcomings in explaining and verifying the used \gls{dnn} models are the key limitation in deploying models in practice. 
Furthermore, existing legal regulations on functional safety~\cite{isotc_22sc_32_iso_2018} and data protection~\cite{goodman_european_2017} are currently extended in upcoming regulations, which will even demand transparency and interpretability of \gls{ai} modules for safe and ethical use.

In this study, we present a novel methodology for post-hoc debugging and testing of computer vision \glspl{dnn}.
Our approach enriches the field of concept embedding analysis with a global-to-local strategy, allowing for locally attributing and testing global concept encodings.
Our local concept-grounded attributions provide a more comprehensive understanding of how global concept encodings are reflected in object detection models. 
We can directly show what is incorporated in the concept encoding, how the concept encoding aligns with the concept semantics, and how the model processes single samples based on the concept usage.


We summarize our contributions as:
\begin{enumerate}
    \item We introduce a novel framework for investigating concept-grounded local feature attributions to debug object detection \glspl{dnn}.
    \item We show how a semantic concept under test is reflected in the general and instance-wise local processing of an \gls{od} model.
    \item We present a qualitative and quantitative comparison of global concept embeddings via locally observable feature importances.
\end{enumerate}



\begin{figure*}[h]
\centerline{\includegraphics[width=\linewidth]{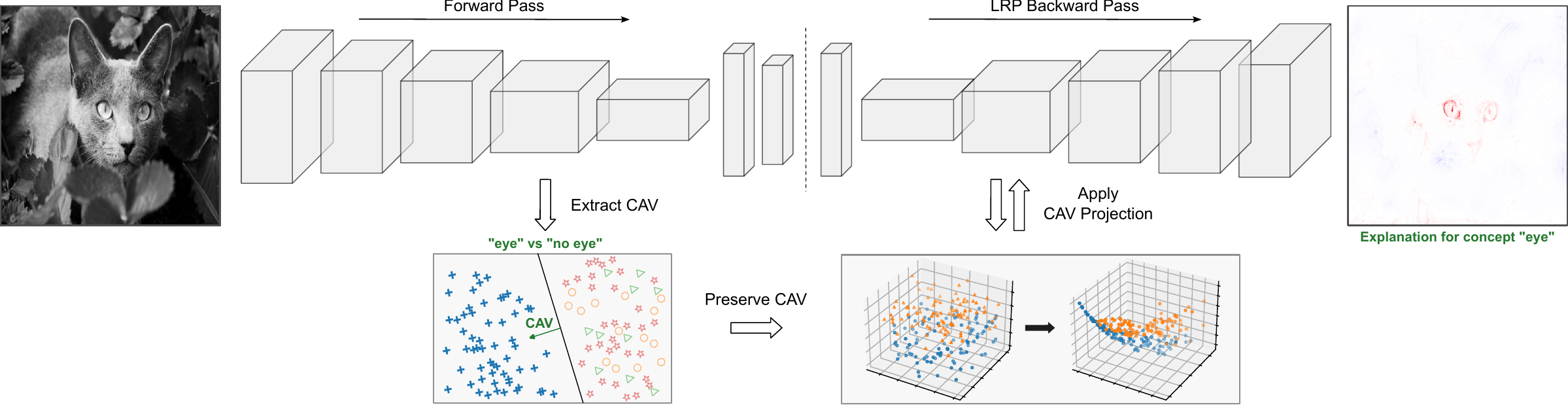}}
\caption{Two-step process of computing \gls{cavlrp} explanations by first extracting a \gls{cav} vector representing the concept in a certain model layer on a concept dataset. Then, the relevance for a sample under test is projected in \gls{cav} direction during the \gls{lrp} backward pass. In this example, \gls{lrp} is initialized to compute an explanation for "cat". The relevance is projected in \gls{cav} direction for the concept "eye" in the specified model layer. }
\label{fig:overview_complete}
\end{figure*}

\section{Related Work}

\subsection{Local Input Attribution}
\emph{Local attribution methods} aim to explain the model processing of a single data sample. Hereby, importance scores for the input units of the data sample are computed concerning a certain model decision.
The computed attribution map signalizes different properties depending on the used \gls{xai} method.
The saliency method~\cite{baehrens_how_2010} provides insights into where the model is most sensitive w.r.t the given input, while increments thereof improve on the predisposition to noise and other properties~\cite{smilkov_smoothgrad_2017,sundararajan_axiomatic_2017,selvaraju_grad-cam_2020}.
Modified gradient backpropagation methods~\cite{fleet_visualizing_2014,bach_pixel-wise_2015,shrikumar_learning_2017} use layer-specific rules to filter and adapt the gradient backward pass of the model w.r.t the output target.
Surrogate methods~\cite{ribeiro_why_2016,lundberg_unified_2017} perform black-box testing by measuring the effect of input alterations on the output.
In this work, \gls{lrp}~\cite{bach_pixel-wise_2015} will be used, as it shows advantageous behavior in sparsity and accuracy of the attribution maps with the focus rather on complete features than individual pixels~\cite{arras_clevr-xai_2022}. 
Compared to the surrogate \gls{xai} methods, it is computationally much more efficient as only one model backward pass is required.


\subsubsection{Layerwise Relevance Propagation}
\gls{lrp}~\cite{bach_pixel-wise_2015} assigns specific rules to each network layer for gradually propagating attribution scores denoted as relevance from the model's output to the model's input. The attribution is hereby distributed from a subsequent layer to the shallower layer based on the neuron activations and their relation to the output target. The propagated relevance score $R_{ij}$ from neuron $j$ to neuron $i$ in the preceding layer -- with $i$ and $j$ denoting the index of the neuron -- is computed using the contribution to the activation of $j$ $z_{ij}$, whereas $z_j$ reflects the sum of contributions to $j$.
\begin{equation}
    R_{ij} = \frac{z_{ij}}{z_j}R_j, \qquad R_i = \sum_j R_{ij}
\end{equation}
The relevance $R_i$ in neuron $i$ is the accumulation of propagated relevance from the neurons of the subsequent layer.
In the ideal case, the sum of attributed relevance in each network layer stays constant, ensuring the property of relevance conservation.
While \gls{lrp} builds the general framework for the backpropagation, several \gls{lrp} rules optimizing different properties have been introduced. Best practices on the selection of rules have been developed for the most common model architectures~\cite{kohlbrenner_towards_2020,pahde_optimizing_2023}.


\subsection{Global Concept Encodings}
\emph{Global \gls{xai} methods} examine the model functionality in general, focusing on dataset-wide assumptions about the learned feature representations and encodings of selected model parts~\cite{holzinger_explainable_2022}.
Concepts build the theoretical construct for finding encodings of semantic information inside the model's latent space.
Instead of using methods for automatically mining concepts~\cite{ghorbani_towards_2019,zhang_invertible_2021,fel2023craft}, which require a human inspector to evaluate the quality of extracted concepts and to assign semantic labels, this work focuses on methods for discovering the model-internal encoding or a semantic concept~\cite{kim_interpretability_2018,fong_net2vec_2018}.

In this work, we want to test a model for the utilization of a specific semantic concept. Hence, we will focus on representations of single semantic concepts.
A concept dataset is required to define the semantic concept and derive the model encoding thereof.
The dataset separates the incorporated samples into two distributions: samples including the specified concept and samples without the concept.
As a well-encoded concept should have its own feature space region in the feature space of at least one of the network layers~\cite{chen_concept_2020}, a linear separation between concept and non-concept data can be assumed.
For this reason, we additionally restrict the concept encodings to linear encodings in the latent space and use the most commonly used methods in this domain: \gls{cav}~\cite{kim_interpretability_2018}, PatCAV~\cite{pahde_patclarc_2022}, and \gls{net2vec}~\cite{fong_net2vec_2018}.



\subsubsection{CAV}
%
%
In \gls{tcav}~\cite{kim_interpretability_2018}, semantic concepts are represented as directional vectors in the feature space, also known as concept activation vectors (\glspl{cav}). The values of single \gls{cav} are equal to the parameters of a binary linear classifier (e.g., SVM) trained on sample activations associated and not associated with the concept. Thus, geometrically, \gls{cav} is a vector orthogonal to the hyperplane separating the concept-related and unrelated regions in the feature space.
%
For given sample set $x$, corresponding concept labels $t \in \{-1, 1\}$, and activation maps $a(x)$ in the selected layer, the optimization task for training the separation hyperplane defining the \gls{cav} $w_{cav}$ is:

\begin{equation}
\begin{aligned}
    &\min |w_{cav}|^2 \\
    \text{s.t.}\quad &t(w_{cav}^T a(x) + b) > 0.
\end{aligned}
\end{equation}

\subsubsection{PatCAV}
\gls{patcav}~\cite{pahde_patclarc_2022} aims to improve on \gls{cav}~\cite{kim_interpretability_2018} by estimating the concept direction as the signal, which is modeled in the decomposition of activations in signal and distractor parts.
The obtained linear concept vector omits distracting activation signals and is therefore more robust to noise in the activations.
Concretely, \gls{patcav} is defined by the correlation between latent activations $a(x)$ of samples $x$ and concept labels $t \in \{0, 1\}$ of the concept-labeled dataset $x, t \in Xh$.



\begin{equation}
     w_{pat} = \frac{cov(a(x),t)}{\sigma_y^2}, \qquad w_{spat} = \sum_{x,t \in Xh} (a(x)- \hat{a}) * (t-\hat{t})
\end{equation}
with $\hat{a}$ denoting the mean activation and $\hat{t}$ being the mean concept label.
By estimating the difference between the mean activations of both data distributions, the simplified version \gls{spatcav} is derived.

\subsubsection{Net2Vec}
A noteworthy distinction of \gls{net2vec}~\cite{fong_net2vec_2018} from CAV-like methods is its ability to perform weak concept localization in the input space.
\gls{net2vec} defines concepts as a linear combination of convolutional filters. In a layer with $K$ filters, the concept is represented by a weight vector $w \in \mathbb{R}^K$.
%
Following this, a linear combination of weights $w$ and the top 0.5\% ($\tau$) of the highest activations $a^{\tau}(x)$ (i.e., denoised activations) is passed through the sigmoid $\sigma(z) = 1/(1+\text{exp}(-z))$ to generate the concept segmentation mask $M(x, w)$:

\begin{equation}
    M(x, w) = \sigma ( \sum_{k=1}^{K} w_k * a^{\tau}_k(x)) 
\end{equation}

This setup is employed for the batch-wise optimization of the \gls{net2vec} weight vector $w$ for a set of samples $x$ and corresponding ground truth concept segmentation masks $L_c(x)$. The optimization objective involves minimizing the binary cross-entropy loss (BCE) between $M(x, w)$ and $L_c(x)$.

\subsection{Combining Local \& Global Approaches}
Multiple works have been proposed for combining concept-based and attribution-based methods.
The Interpretable Basis Decomposition framework~\cite{ferrari_interpretable_2018} defines a concept-based decomposition in a linearly decomposable layer w.r.t the output, where the concept information can be locally explained using \gls{gradcam}~\cite{selvaraju_grad-cam_2020}. 
Similarly, in \cite{brocki_concept_2019}, a concept vector is trained in the latent space between encoder and decoder of a \gls{vae} before computing and locally attributing the dot product between \gls{cav} and the encoder output.

ProtoPNet~\cite{chen_this_2019} builds up prototypes to compare features of new samples to prototypical class features. Local explanations for relating new samples to the prototypes are computed using GradCAM~\cite{selvaraju_grad-cam_2020} and highlight regions in the input sample that resemble the respective prototype.

Using the Kernel trick, Crabbé and van der Schaar~\cite{crabbe_concept_2022} apply a non-linear separation of concept and non-concept activations. Concepts are thereby represented as \glspl{car}. The concept density is hereby used for measuring the relation of samples regarding a specified concept. Local attribution scores are derived based on a density value~\cite{crabbe_concept_2022} between a sample activation and the \gls{car}.

In \gls{crp}~\cite{achtibat_where_2022}, a concept-based extension to \gls{lrp} is implemented in the form of a filter-based masking. The relevance pass is guided through a single convolutional filter in the latent space to explain an output w.r.t this filter encoding. Human inspection is needed to assign a semantic concept label to explanations of the same filter.
An extension to object detection models is provided in~\cite{dreyer_revealing_2022}.

\cite{vielhaben_multi-dimensional_2023} defines concepts as multi-dimensional subspaces, which can be locally explained with CAM-like upsampling~\cite{selvaraju_grad-cam_2020}.

Further methods seek to mine concepts by disentangling the latent space into subspaces, each encoding an individual concept. While in~\cite{chormai_disentangled_2022} subspaces are mined based on relevance and disentanglement, the criteria of disjunction and identifiability are used in~\cite{leemann_when_2023}. 

While many of these approaches require specialized network architectures or build upon concept discovery, we purely base our approach on post-hoc feature-space attributions and pre-defined semantic concepts from a concept dataset.

\section{Local Concept-based Attributions}



\begin{figure*}[h]
\centerline{\includegraphics[width=0.9\linewidth]{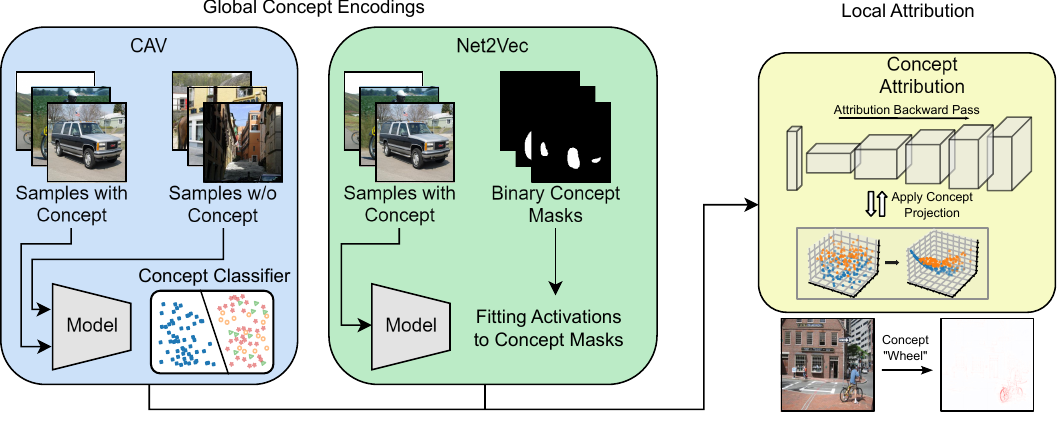}}
\caption{Overview of our \gls{cavlrp} framework consisting of (1) multiple options for computing linear concept encodings (left) and (2) manipulating the attribution backward pass for deriving local concept-based explanations (right). }
\label{fig:approach_complete}
\end{figure*}

\subsection{Global-to-local Concept Attribution}



Our proposed framework continues the promising direction in \gls{xai} of combining local and global explainability approaches.
While local \gls{xai} methods can describe model functionality on single data samples, global methods can describe semantic model behavior~\cite{holzinger_explainable_2022}. 
We use global encodings of single concepts and adapt the attribution backward pass of local attribution methods towards the concept encoding to compute concept-based attributions for single data samples and corresponding model predictions. 

Choosing a specific layer $l$ of a \gls{dnn}, the model function $f:\mathbb{R}^{n}\rightarrow\mathbb{R}^{m}$ can be defined in the form $f(x) = h(g(x))$ with $g:\mathbb{R}^{n}\rightarrow\mathbb{R}^{k}$ specifying the model function prior to layer $l$ with $\mathbb{R}^k$ describing the latent space between $g$ and $h$, and $h:\mathbb{R}^{k}\rightarrow\mathbb{R}^{m}$ specifying the model part till the output.

In the first step (1) of \gls{cavlrp}, a concept representation in the form of a linear vector $v_{CAV}^l$ is computed for a single semantic concept using a dataset with adequate labeling for this concept.
The linear vector $v_{CAV}^l$ is derived from the activations of feature layer $l$ and shall describe the specified concept direction in the latent space. Any global xAI method providing a linear concept encoding, like \gls{cav}, \gls{patcav}, or \gls{net2vec}, is applicable for computing the concept vector.
For step (2), the concept-based attribution for a single sample and an arbitrary target initialization regarding the model output is computed. 
We choose to use \gls{lrp} for its sparse and pixel-exact attributions, but any other local backpropagation method can be used respectively.
With $R^h(x)$ denoting the relevance at layer $l$ for an input $x$,
the relevance $R^h(x)$ is linearly projected in the direction of the concept $v_{CAV}^l$:
\begin{equation}
    R_{CAV}^h = R^h \cdot v_{CAV}^l.
\end{equation}
For deriving the concept attribution in input space in step (3), the projected relevance $R_{CAV}^h$ is then propagated back through the model part $g$ to the input layer as before, following the rules of \gls{lrp}~\cite{bach_pixel-wise_2015}.

Note that the \gls{lrp} constraint of relevance conservation in adjacent layers does not hold due to the included projection. This is thus reasonable, as the projection of relevance in \gls{cav} direction can intuitively be seen as a filter for removing non-concept information in the intermediate relevance.

\subsection{Applicability in Object Detection}
Profound evaluations on configuring the \gls{lrp} rule composites and initializing the backward pass have been developed for certain standard classification models~\cite{kohlbrenner_towards_2020,motzkus_measurably_2022,pahde_optimizing_2023}. However, there is no best practice for applying \gls{lrp} to object detection models.
For the rule composition on the used \gls{od} models, we follow the general principles for classification models of using the \gls{lrp}-$\epsilon$ rule for deep layers while attributing the convolution layers in the backbone with the \gls{lrp}-$\alpha 1 \beta 0$ rule.
Canonization is applied for merging batch normalization layers into adjacent linear layers.

For initializing the attribution backward pass, we condition the target on the class probabilities of the detected objects to obtain class-based feature scores rather than explaining the regression problem of localizing the object.
While~\cite{dreyer_revealing_2022} restrict their method to single detections and handle each detected object individually as a classification decision, \cite{karasmanoglou_heatmap-based_2022}~initialize the back-propagation specifying single object classes, so that the relevance for multiple objects of the same class is computed simultaneously.
Our approach extends this idea by providing multiple options based on the tested objective.
For testing whether and how a concept has been used in the model processing of a complete sample, the backward pass is initialized with the positive activation values in the direct model output. Instead of performing a \gls{nms} filtering, the output is clipped to positive activations only and scaled to a range of $[0,1]$ for numerical stability. 
Clipping to positive activations inhibits negative attribution signals to superimpose the positive attribution of the concept signal.
Alternatively, the \gls{nms}-filtered detections of the model can be tested for the usage of a concept by constructing a binary mask, which incorporates the selection of anchor boxes and selected classes individually or in conjunction.
With the approach, it can thus be evaluated how and whether the concept has been used for (a) the processing of the complete input, (b) a selection of objects, or (c) a single detection made by the model.

\begin{figure*}[h]
\centerline{\includegraphics[width=0.7\linewidth]{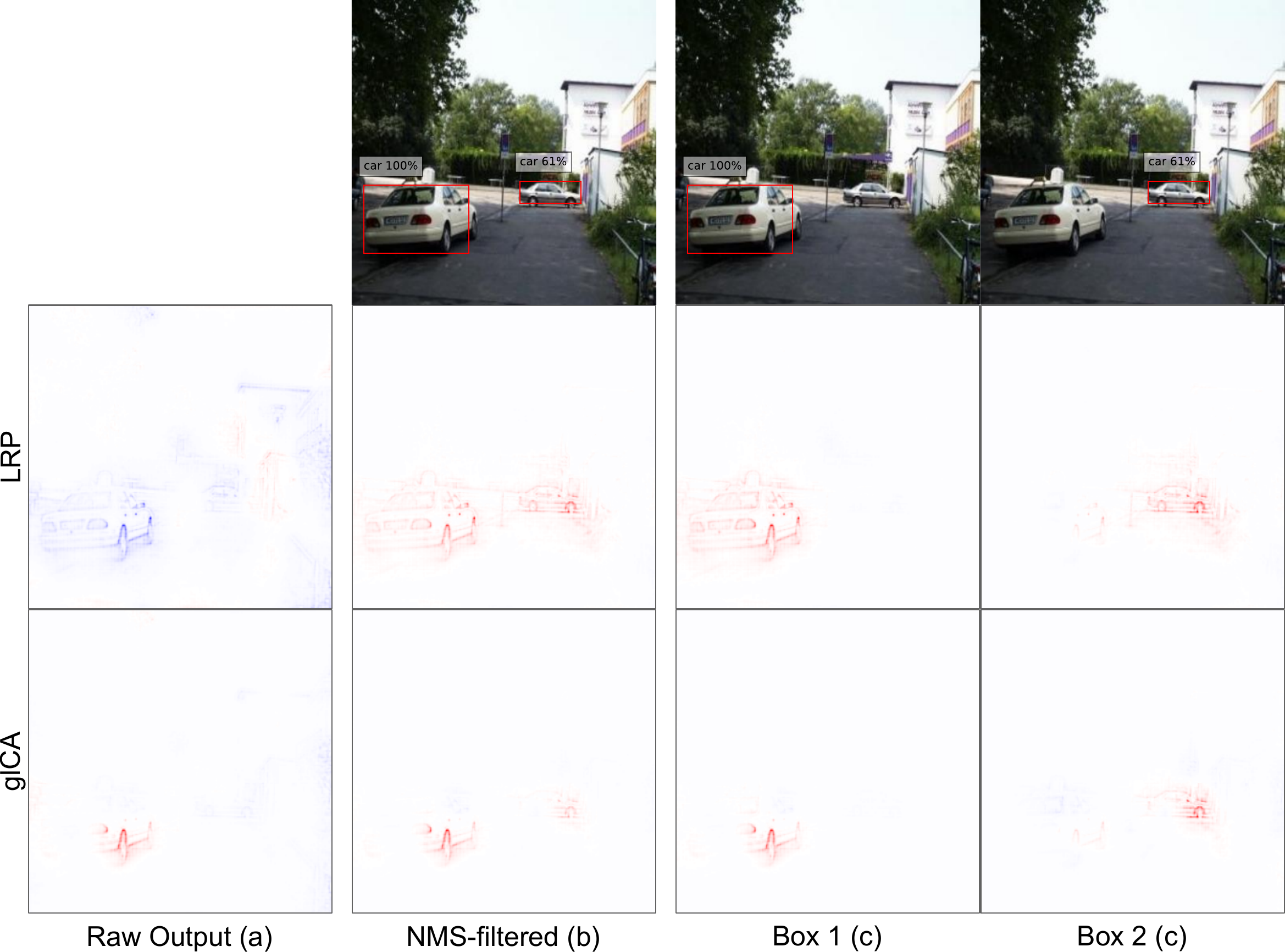}}
\caption{Visual example of the different options (A-D) of using \gls{lrp} and \gls{cavlrp} for testing an SSD model for the concept "wheel" in layer \texttt{feat.19}. }
\label{fig:case_overview}
\end{figure*}

\section{Quantification measures}
\label{sec:metrics}


For our experiments, we align our evaluations with current approaches in \gls{xai} quantification~\cite{hedstrom_quantus_2023} for classification models, where desirable properties of local attribution methods are quantitatively tested with regard to the class output. 
We focus on the main qualities of the \emph{localization} capability and the \emph{faithfulness} testing, which we both extend for concept-based attributions.
The localization capabilities of an \gls{xai} method are particularly important, as especially local attribution approaches mainly describe where important features are located inside an image.
The faithfulness is investigated as \gls{xai} methods are only valid and useful if they describe the actual model behavior.


\subsection{Concept Localization}


\subsubsection{Concept Attribution Localization}
As \gls{cavlrp} can be used for explaining the detection of a certain concept within an input, we want to measure the localization capabilities in the explanations for Option (a) of our approach.
We, therefore, use an adapted version of the attribution localization metric~\cite{kohlbrenner_towards_2020}, which is implemented in the Quantus toolbox~\cite{hedstrom_quantus_2023}, that measures the ratio of positively attributed relevance within a binary class mask to the overall positive relevance.
This metric can also be applied on the concept level when concept mask annotations are provided, as in the Broden~\cite{bau_network_2017} dataset.
The concept attribution localization $\mu_c$ regarding the concept $c$ is defined as
\begin{equation}
    \mu_c = \frac{CR_{in}^+}{CR_{total}^+}
\end{equation}
with $CR_{in}^+$ denoting the positive \gls{cavlrp} attributions regarding concept $c$, which fall within the annotated concept area and $CR_{total}^+$ denoting the overall positive attributions of the \gls{cavlrp} explanation regarding concept $c$.
The concept attribution localization measure assumes that the model uses the local concept information in the input to detect this concept.

\subsection{Faithfulness Testing}
Testing the faithfulness of our attribution approach to the model's detections needs to be split into two steps.
In the first step, similar to input perturbation testing~\cite{samek_evaluating_2017}, we observe whether the concept attribution is faithful to the model concerning a consistent ranking of the input pixels.
As the concept attribution map provides a pixel-level ranking based on the pixel importance for a selected detection and concept direction, removing the most important pixels should harm the detection most. 
By removing an increasing number of pixels following this ranking, the concept information used in the image is expected to be discarded first.
During the process, the class score of a single detected object is observed. 
If the scoring decreases, the attribution reflects the model's use of the attributed pixels, which reflect the overlap of object class and concept information.

The second step observes the relation between concept information and detected objects.
Hereby, the attribution share in concept direction from the overall attribution is monitored during the input perturbation.
As the concept information in the input should be discarded first, we expect the attribution share to attenuate in correlation to the predicted object scores.



\section{Results}

\subsection{Experimental Setting}
For our evaluation, a split of the Broden~\cite{bau_network_2017} dataset, which contains concept annotations for, e.g., parts or materials, is used. 
We evaluate our method on two commonly used object detection models, an \gls{ssd} with a VGG~\cite{simonyan_deep_2014} backbone and the FasterRCNN with a ResNet~\cite{he_deep_2016} backbone. Since both backbones are also commonly used in classification models, the results can be transferred to the easier task of classification without further constraints. 
Both tested models are pre-trained on the COCO~\cite{lin_microsoft_2014} dataset and not explicitly trained on Broden, but they nevertheless perform well in detecting COCO-related objects in the Broden samples.
The global concept encodings are trained on samples from the Broden training set. While \gls{net2vec} naturally provides a one-dimensional vector, covering the number of filters in a layer, \glspl{cav} can be represented in the three-dimensional space.
We obtain the representations for \gls{cav} and \gls{patcav} by averaging across the spatial dimensions to obtain a one-dimensional concept vector as in \cite{graziani_concept_2020}, which yields improvements in computational efficiency.
For computing local attributions with \gls{lrp}, the Zennit toolkit~\cite{anders_software_2021} is used.

\begin{figure*}[h]
\centerline{\includegraphics[width=\linewidth]{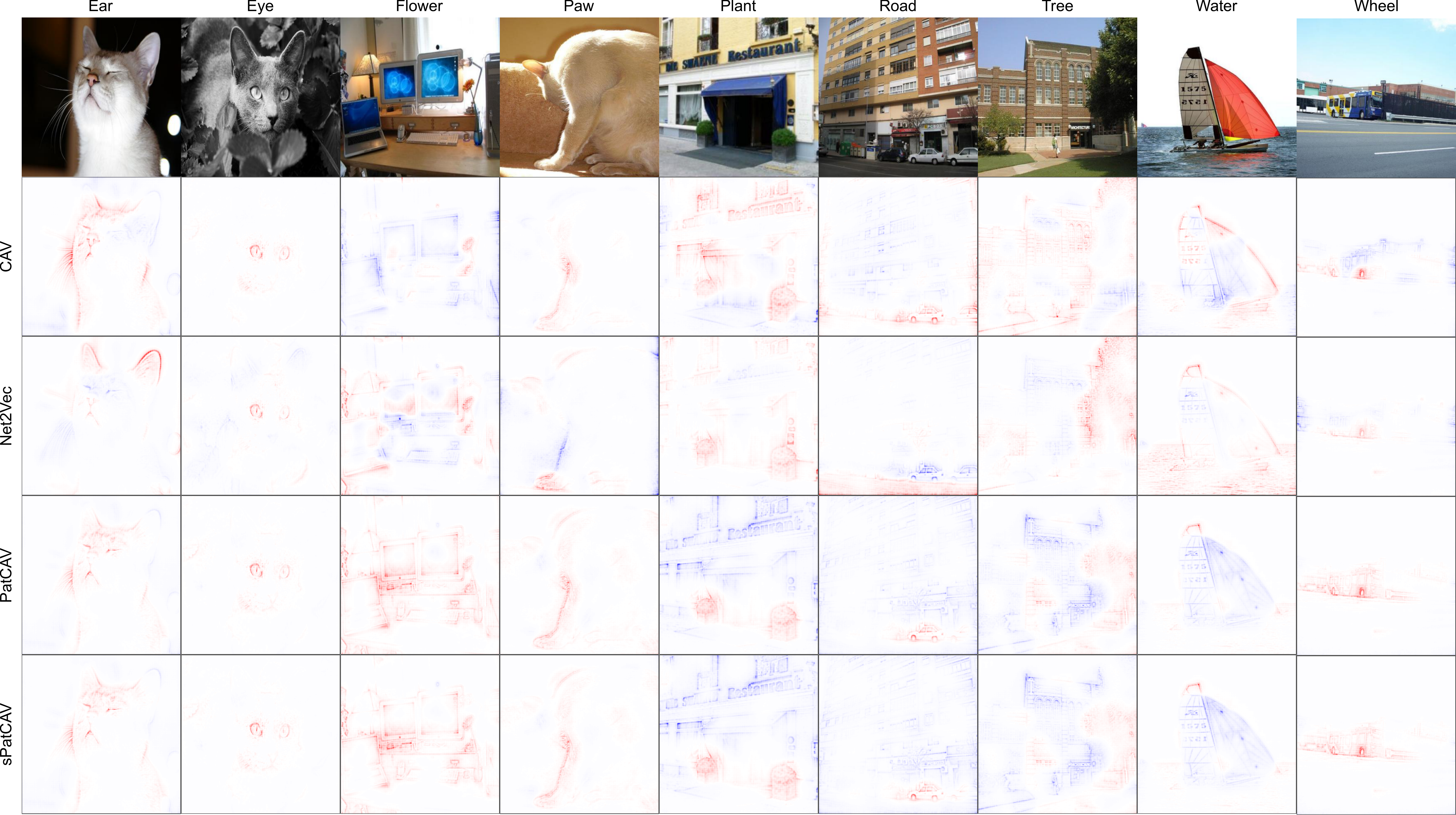}}
\caption{\gls{cavlrp} explanations with Option a for multiple concepts w.r.t. \glspl{cav} in the best suitable layer \texttt{feat.\{17,19,21\}} of the SSD model. }
\label{fig:ssd_example_all}
\end{figure*}








\subsection{Local Concept Attribution}




Testing an object detection model for the usage of a certain semantic concept, multiple options w.r.t the tested object(s) can be inspected.
Figure~\ref{fig:case_overview} illustrates the \gls{cavlrp} explanations for the options of general concept usage [a\&b] and detection-based concept usage [c]. 
While the \gls{lrp} explanation highlights the full object features, a clear concentration on the concept \enquote{wheel} can be observed for all options of our concept-based explanations.
This selective attribution underscores the ability of the concept encoding to model the concept direction in the model's latent space and the ability of \gls{cavlrp} to attribute the concept usage to the input accurately.
Figure~\ref{fig:ssd_example_all} shows the applicability of our approach with multiple concept encodings to a random selection of different concepts from the Broden dataset.

\begin{figure*}[h]
\centerline{\includegraphics[width=0.9\linewidth]{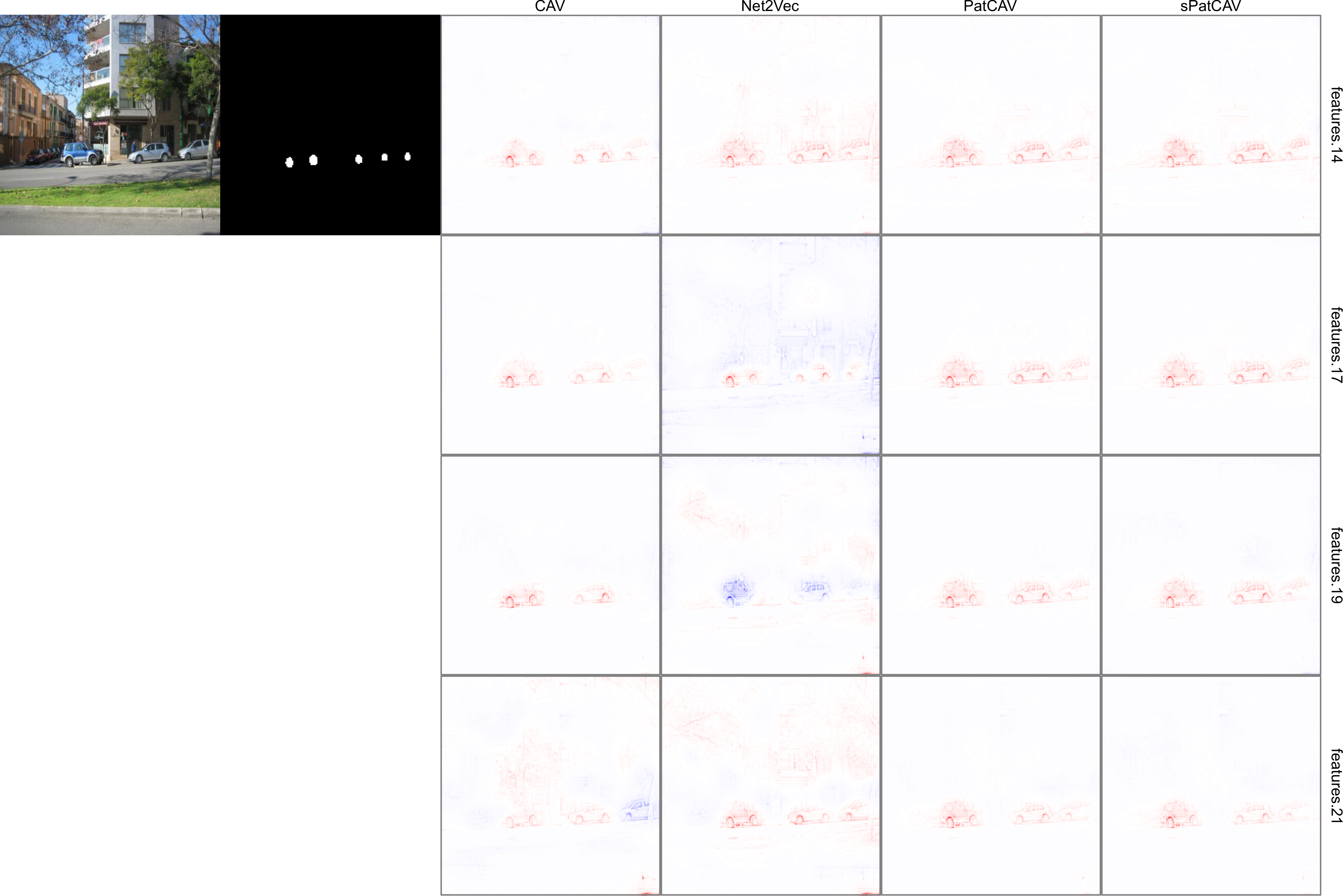}}
\caption{Visual comparison of the explanations for different global concept encodings for multiple layers of the SSD model. }
\label{fig:layer_examples}
\end{figure*}

The quality of the concept attribution maps in terms of highlighting the concept is naturally determined by the global concept encoding and the forward processing of the selected sample.
Qualitatively convincing attributions are only provided if the concept can be linearly encoded in the latent space of the selected layer and if this concept encoding is used by the model for the detections in the selected sample.
Figure~\ref{fig:layer_examples} shows the respective \gls{cavlrp} explanations (Option a) for concept encodings using \gls{cav}, \gls{net2vec} and \gls{patcav} in multiple layers of the \gls{ssd} model.
The explanations significantly vary in the distribution of attributed pixels.
While the explanations for \gls{cav} and \gls{net2vec} depend on the selected layer, \gls{patcav} seems to be more stable, as the explanations only slightly change over the layers.
However, for the right layer choice, the explanations for \gls{cav} and \gls{net2vec} are more targeted to the concrete concept information.
\gls{patcav} also attributes non-concept information and provides more scattered explanations.

Discriminating or correlated features regarding the concept do not necessarily focus on the concept but can be prone to general correlations in the two provided data distributions.
Optimizing for concept masks is more targeted to where the concept to learn is located but cannot guarantee whether the concept encoding reflects the concept semantically.

\subsection{Evaluating Concept Usage}
\begin{figure*}[h]
\centerline{\includegraphics[width=0.8\linewidth]{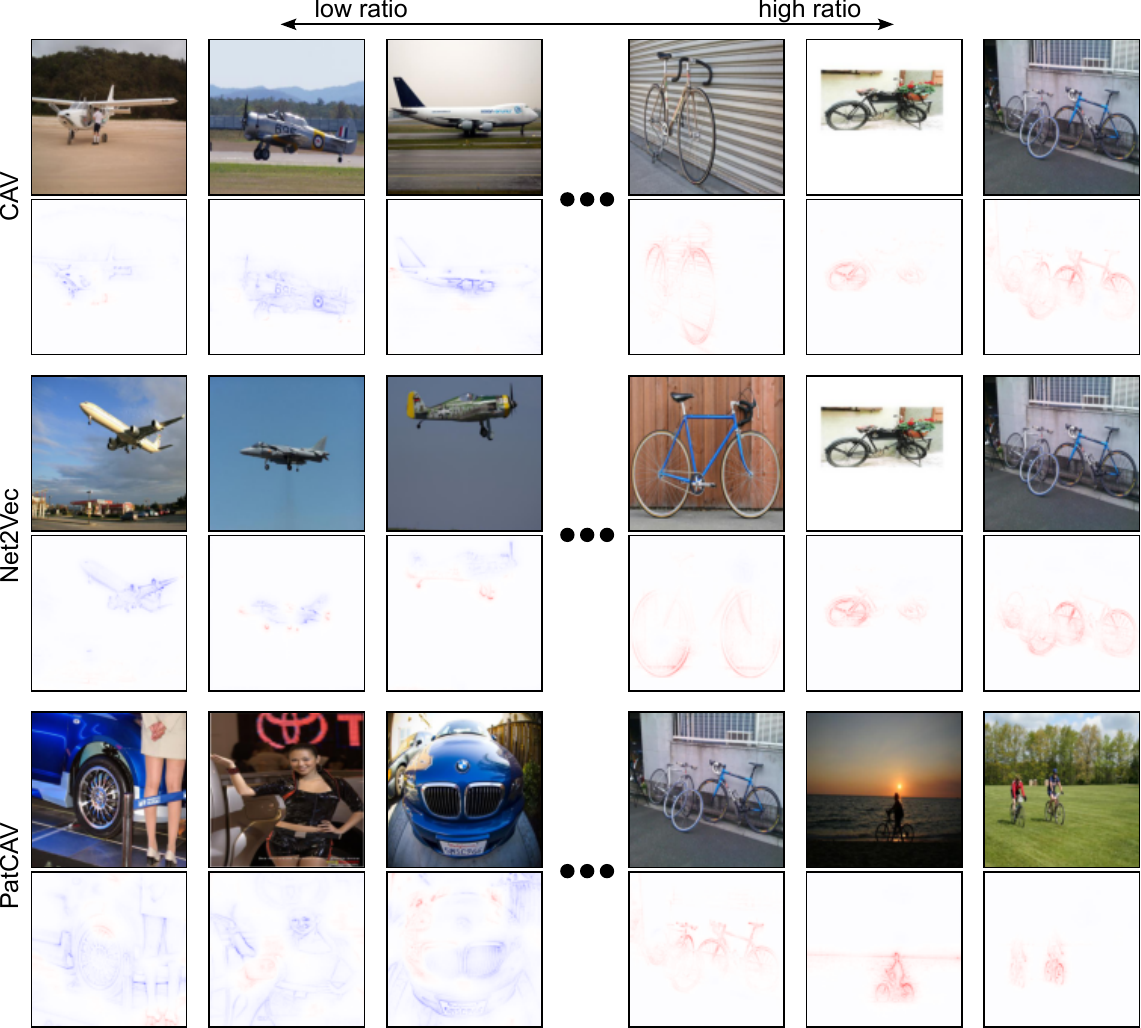}}
\caption{Samples and their explanations sorted by concept usage for concept \enquote{wheel} for different global concept encodings in layer \texttt{feat.21}. Left: Low concept usage, Right: High concept usage. }
\label{fig:usage19}
\end{figure*}

Comparing the latent attribution with and without the concept-based projection, a ratio of concept importance in processing a sample can be computed.
Sorting the input samples and respective explanations by concept usage (Figure~\ref{fig:usage19}) shows which concept appearances are reflected best and which concept appearances are not encoded. 
We thereby see a tendency of \gls{net2vec} rather to encode larger concept appearances due to the concept mask resolution in the training. 
Generally, a semantic correlation between samples with a similar ratio of concept usage can be observed.
Hereby, samples with similar concept usage resemble each other with regard to the size and the context of the concept. 
For the concept \enquote{wheel}, the contextual information incorporates the orientation and the vehicle type it is part of. 
Generalizing this finding towards global concept vectors, their capability of encoding a concept semantically seems questionable, as the trained encoding rather learns single illustrations of the concept, which are consistent in size, orientation, and context, than fully representing the semantic concept.

\begin{figure*}[h]
\centerline{\includegraphics[width=0.75\linewidth]{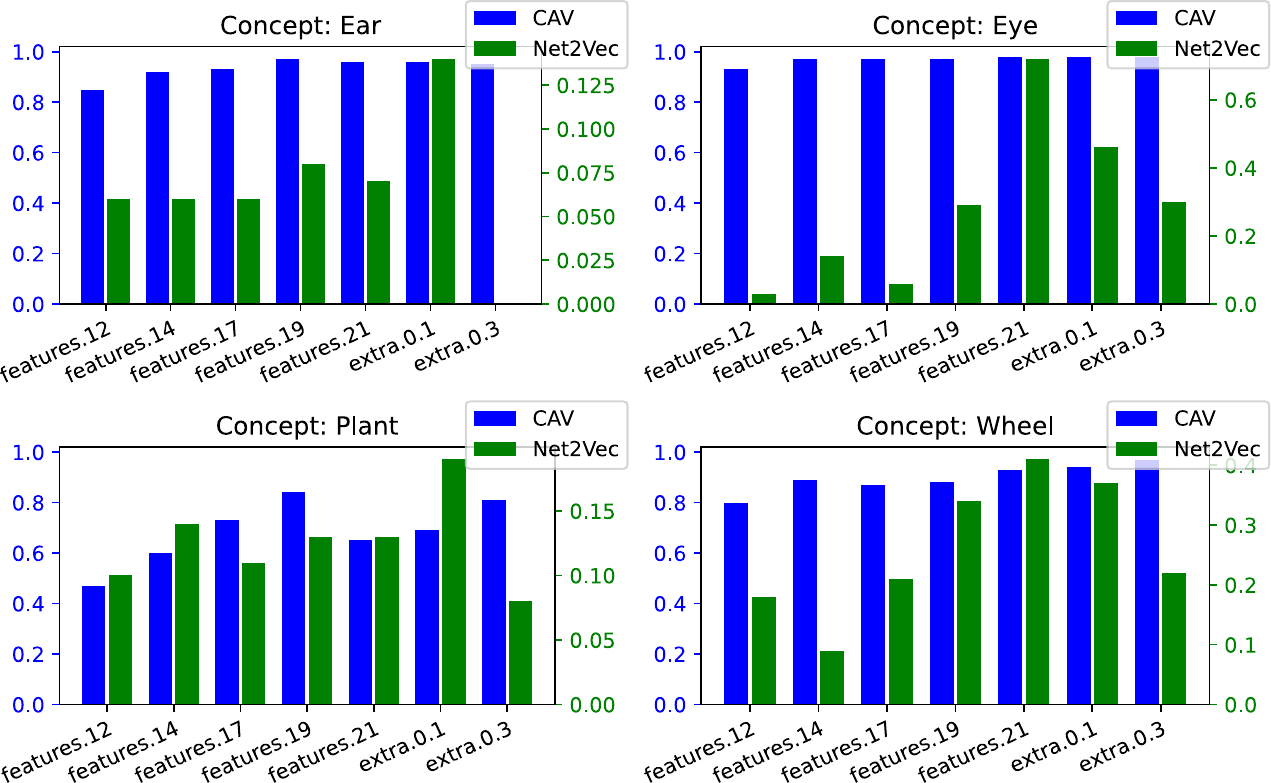}}
\caption{Performance scores from the training of the concept vectors. For \gls{cav}, the classification accuracy on hold-out data samples is reported, while for \gls{net2vec}, the IoU score on hold-out samples is reported.}
\label{fig:training_scores}
\end{figure*}

\subsection{Localization Quantification}

\begin{figure*}[h]
\centerline{\includegraphics[width=0.75\linewidth]{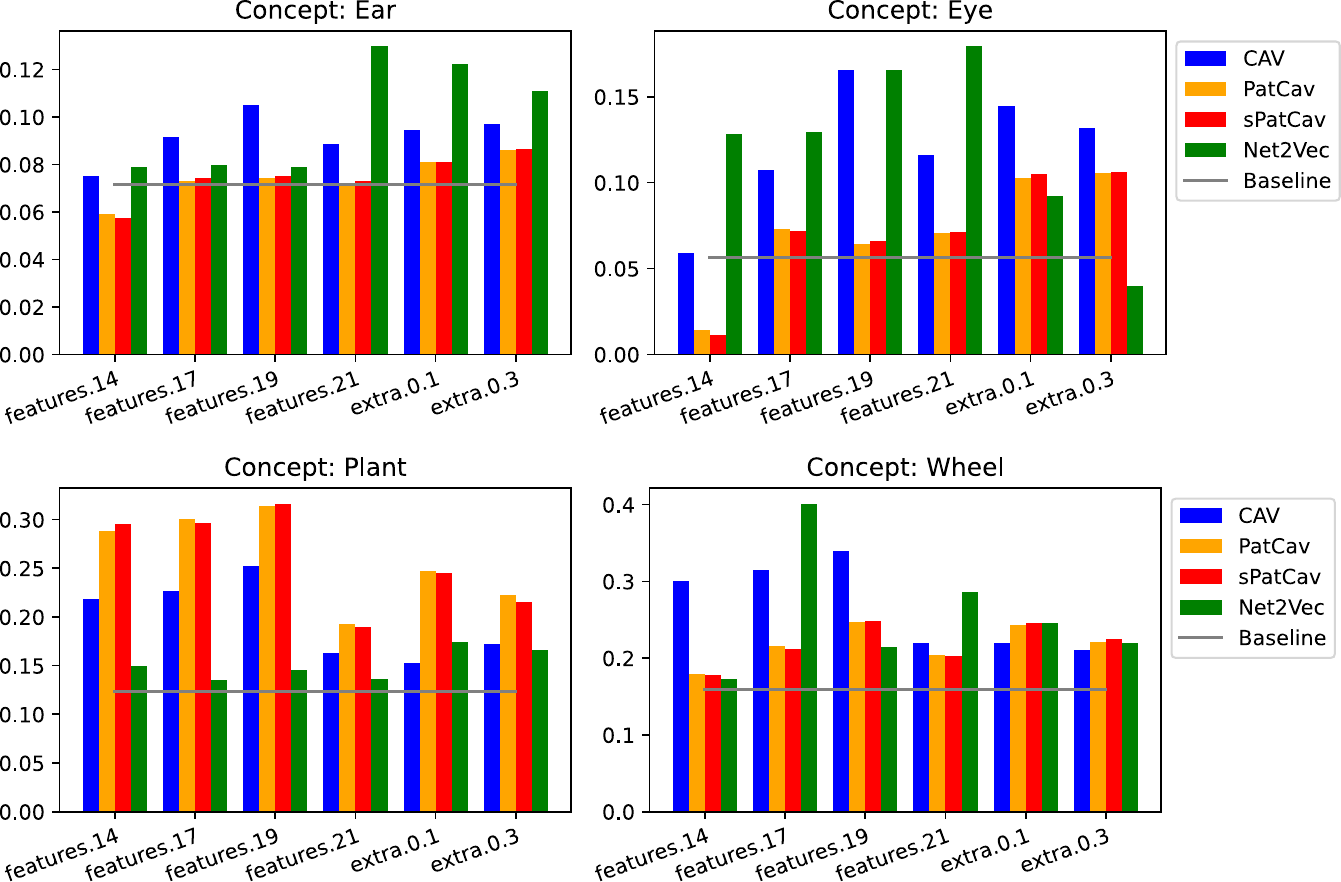}}
\caption{Quantitative comparison of the concept attribution localization over multiple convolution layers in the backbone (features) and extra layers of the SSD model for the concepts \enquote{ear}, \enquote{eye}, \enquote{plant}, and \enquote{wheel} (higher is better). }
\label{fig:localization_combined_all}
\end{figure*}

Comparing the attribution localization results in Figure~\ref{fig:localization_combined_all} to the relative performance scores from Figure~\ref{fig:training_scores}, a clear correlation between the localization performance and the training scores can be seen for \gls{net2vec}, while the correlation is not reflected for \gls{cav}.
Nonetheless, the localization scores for the different global encodings agree in the range of layers with the best scores, which indicates that the concept is likely encoded in these most properly.



For most concepts, \gls{net2vec} achieves the best localization score for the best fitting layer.
This finding is not surprising as the \gls{net2vec} training already involves the optimization for concept masks reflecting localized information, whereas the other global encodings are not optimized using local alignment.
While the localization performance of \gls{patcav} and \gls{spatcav} is generally similar, their scores are significantly lower in most layers. 
However, for single concept \enquote{plant} they seem to find a better representation of the concept.
In the general case, their attribution maps tend to highlight all concept-related features, including other parts of the object and background, when there is a correlation to the concept.
A clearer distinction between the features in the training data might be needed to improve the \gls{patcav} encoding in this regard.


A general finding cannot be given as to which concept embedding works best. 
\gls{net2vec} is, per definition, trained for location-constraint information and is most accurate for optimized parameters.
\gls{cav} and \gls{patcav} are trained for separating distributions.
The used information is hereby not directly describing the concept, but also smaller features of the concept or surrounding information that correlates with the concept.
While \gls{patcav} incorporates surrounding information for most concepts, it surprisingly performs best for single concepts (like~\enquote{plant}, Figure~\ref{fig:localization_combined_all}), where \gls{cav} and \gls{net2vec} fail to find a good encoding.

\subsection{Faithfulness Evaluation}

\begin{figure*}[h]
\centerline{\includegraphics[width=\linewidth]{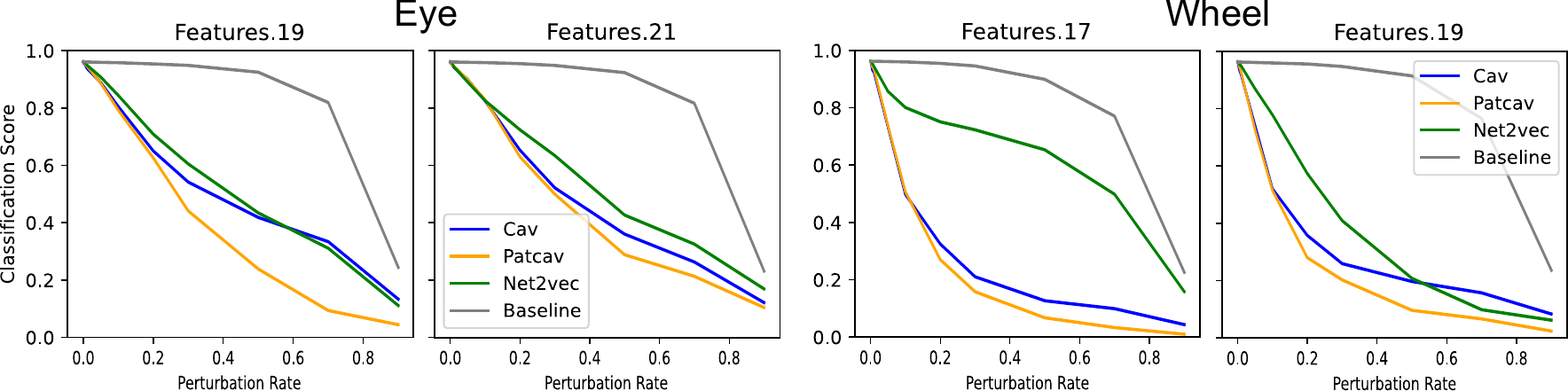}}
\caption{Quantitative comparison of the input perturbation testing over multiple layers of the SSD model for the concepts "eye" and "wheel" (lower is better). }
\label{fig:faithfulness_combined_all}
\end{figure*}

The alignment between \gls{cavlrp} explanations and the actual model processing is examined in a faithfulness evaluation.
The first faithfulness assessment compares the pixel importance in the \gls{cavlrp} explanation with the model's response during the input perturbation testing.
As can be seen from Figure~\ref{fig:faithfulness_combined_all}, our \gls{cavlrp} explanations perform much better in assigning importance scores than a random pixel selection, which reflects a general alignment of the assigned pixel importance to the model processing.
Especially \gls{patcav} yields a strong alignment to the class-sensitivity. 
Observing the attribution maps, this phenomenon may stem from \gls{patcav} encoding more contextual information, potentially shifting the focus from purely conceptual features to generally relevant features.
A stronger response in the input perturbation testing is thus expected.
While \gls{net2vec} shows a similar drop in the model score for the first perturbation steps, the scores are noticeably higher, particularly in layer \texttt{feat.17}. 
This can be explained by the method not finding a suitable encoding in this layer, while in other layers, \gls{net2vec} attributes non-concept features negatively. 
As a result, pixels lacking informative content are perturbed prior to the removal of those associated with objects but not the concept.
The object-related class score is thus higher.



\begin{figure*}[h]
\centerline{\includegraphics[width=0.8\linewidth]{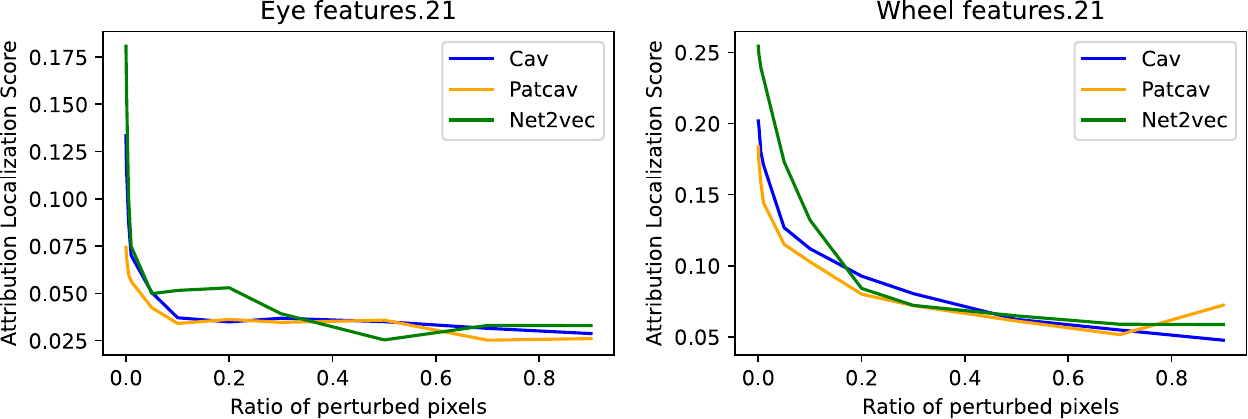}}
\caption{Attribution Localization scores during the input perturbation testing (lower is better). Removing pixel information via the importance ranking from the \gls{cavlrp} explanation has a significant impact on lowering the attribution localization scores indicating the faithfulness of the explanation to the model-internal concept encoding. The results refer to encodings in layer \texttt{feat.21} of the SSD model. }
\label{fig:faithfulness_locscore}
\end{figure*}

\begin{figure*}[h]
\centerline{\includegraphics[width=0.8\linewidth]{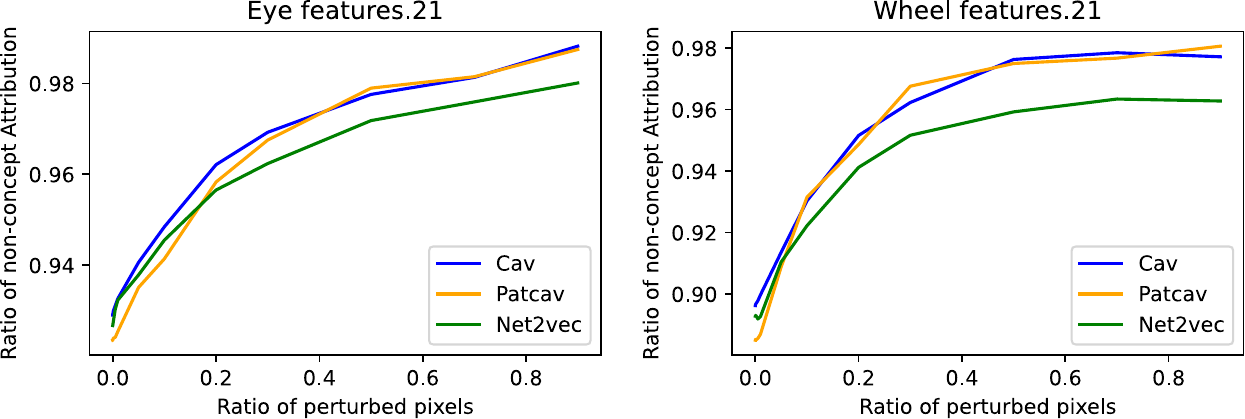}}
\caption{Ratio of latent space attribution not related to the encoded concept during the input perturbation testing. A score of 1 denotes zero attribution in the concept direction (higher is better). The scores converging to $1.0$ indicate the removal of latent space attribution in the direction of the concept encoding. The results refer to encodings in layer \texttt{feat.21} of the SSD model. }
\label{fig:faithfulness_diffratio}
\end{figure*}

For testing whether the \gls{cavlrp} explanations reflect the concept's influence on the processing of the model for single detections, we further test the concept-related attribution during the input perturbation.
Hereby, we measure how the attribution localization score and the ratio of concept-projected to actual attribution change during the input perturbation (Figures~\ref{fig:faithfulness_locscore} and \ref{fig:faithfulness_diffratio}).
Both measures show a strong score decrease/increase during the first perturbation steps, which hints at a strong correlation between the explanations and the model's concept usage.
Thus, we show that the explanation is faithful to the concept encoding in the latent space and that the explanation is faithful to the overall usage of the concept in the model's detections.


\section{Testing for erroneous feature correlation}

\begin{figure*}[h]
\centerline{\includegraphics[width=\linewidth]{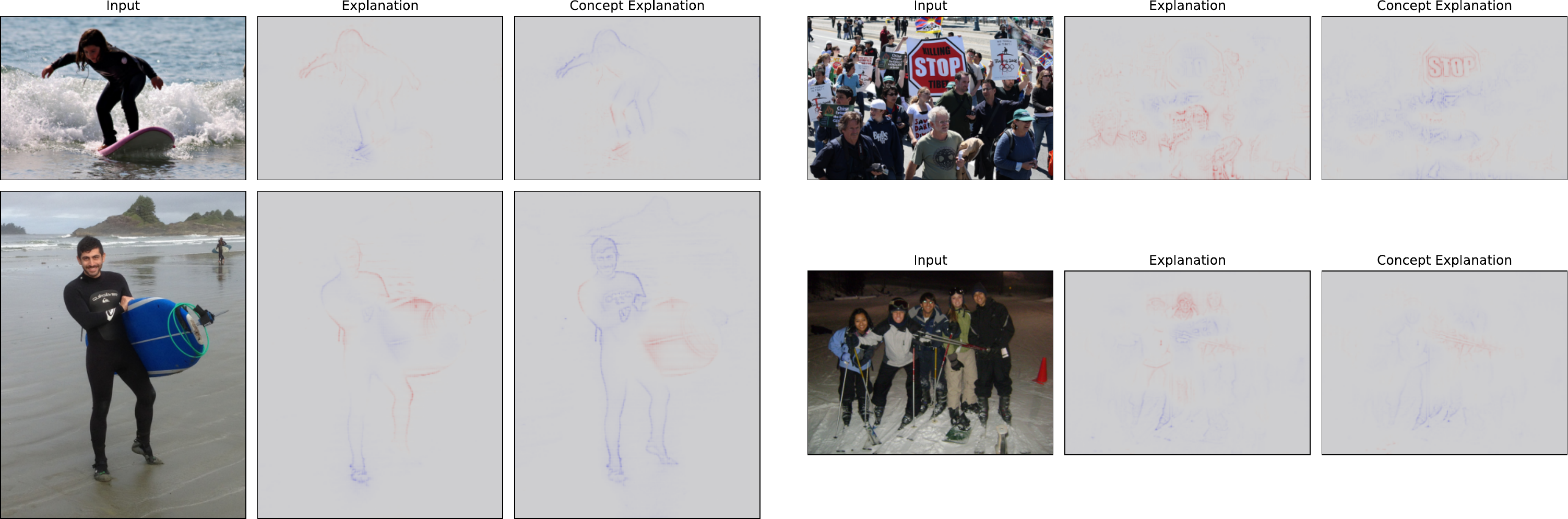}}
\caption{Comparison of the attribution maps of standard \gls{lrp} and \gls{cavlrp} w.r.t concept "surfboard" (left), "stop sign" (top right) and "skis" (bottom right) for a detected instance of class "person" in layer \texttt{feat.21}. In the \gls{cavlrp} explanation, the positive attribution clearly focuses on the tested concept exposing the usage of this information for the detection. } 
\label{fig:coco_spurious}
\end{figure*}

We want to test an object detection model for the usage of semantically unnecessary features in its detections on the commonly used dataset COCO~\cite{lin_microsoft_2014}.
As can already be seen from the class composition, the COCO dataset contains a substantial amount of samples with multiple classes included. 
The cumulated connection of samples from two classes may induce spurious correlations.
We explore the influence of class-related concept encodings on the detections of class "person".
We therefore train a \gls{net2vec} concept encoding for multiple concepts and test the importance of the concept on single detections of class \enquote{person} on evaluation samples.
The derived \gls{cavlrp} explanations show clearly visible positive attributions to concept features of the tested concepts indicating a distinct correlation and utilization of these features for the detection.
Although the concept features are semantically not necessary for detecting the class object, we can find these contextual correlations for several concepts (Figure~\ref{fig:coco_spurious}).
The results imply a need to investigate the suppression of spurious feature correlations further.


\section{Conclusion}

With our proposed \gls{cavlrp} framework, we extend the current \gls{xai} research in combining global concept embeddings with local attributions to provide a better understanding of global concept representations on single data points.
Further applying concept attributions to two object detection models, we show how to test single detections as well as the general model processing for a global concept.
Our qualitative and quantitative evaluation assesses the differences in multiple global concept encoding methods, reflecting their relation to a concept's semantic meaning as well as the relation to the processing of a single sample.
Hereby, we emphasize the discrepancy between the semantic concept and the information encoded in the linear concept encoding.
While performance metrics in distinguishing between distributions do not reflect the quality of the encoding in a selected layer, our concept attribution score can help to find proper concept encodings.
Our use case on COCO class correlations reveals the model usage of unnecessary or even misleading concepts in detections of class \enquote{person}, soliciting additional model debugging before applying detection models in practice.
We highlight the importance of local concept-based explanations for assessing the quality of concept encodings and encourage further research towards the feature variety belonging to a semantic concept.

\begin{credits}
\subsubsection{\ackname} The research leading to these results is funded by the German Federal Ministry for Economic Affairs and Climate Action within the project \enquote{KI Wissen -- Entwicklung von Methoden für die Einbindung von Wissen in maschinelles Lernen}. The authors would like to thank the consortium for the successful cooperation.

\subsubsection{\discintname}
The authors have no competing interests to declare that are
relevant to the content of this article.
\end{credits}

%
%
%
\bibliographystyle{splncs04}
\bibliography{xai}

\clearpage
\appendix
\section{Selection of Concepts}
In this work, concepts from the Broden parts annotation have been used. Hereby, concepts have been selected based on the following criteria:
\begin{itemize}
\item \textbf{Representativity}: Due to the training of the CAVs and our quantitative evaluation, concepts with a high number of validation data points are favored.
\item \textbf{CAV Precondition}: As a CAV needs to be properly trained for our approach, only concepts with a binary classification (concept/no concept) accuracy of over 85\% are considered.
\item \textbf{Diversity}: Some concepts in the part annotations can be directly linked to object instances (e.g., \textit{eye} or \textit{wheel}), while others are not physically linked to instances (e.g., \textit{clouds} or \textit{water}). The used concepts have been chosen in order to cover both cases.
\item \textbf{Randomness}: The used concepts have been chosen randomly from the remaining candidates to increase the variability.
\end{itemize}

\section{Implementation Details \& Color Coding}
For computing relevance scores with \gls{lrp}, composite rule strategies have been used as suggested in \cite{kohlbrenner_towards_2020}.
For the SSD, the $\epsilon$-rule has been used for the model heads, while the $\alpha1\beta0$-rule was used for lower layers. 
Additionally, the VGG Canonizer from zennit~\cite{anders_software_2021} has been applied together, while also batch normalization layers have been merged into adjacent linear layers.
Similarly, batch normalization layers were merged for the FasterRCNN, and the ResNet canonizers from zennit were applied.
A similar composite consisting of applying the $\epsilon$-rule to the heads and the FPN and applying the $\alpha1\beta0$-rule to lower layers was used.

As object detectors typically have a complex post-processing, including the NMS filter, the processing and initialization of relevance needs to be shifted to the previous network layer output. A mapping functionality is therefor implemented, mapping the explanation target back to the output representation consisting of all proposals and class probabilities.

For the colorization of the explanations, the zennit~\cite{anders_software_2021} toolbox is used. The red color indicates a positive attribution stating that the respective pixels have been used in favor of the selected class. The blue color indicates negative attribution stating that these pixels do not support the selected class.

\section{Comparison to CRP}

The comparison of our \gls{cavlrp} framework to \gls{crp}~\cite{achtibat_where_2022} holds difficulties in finding and assigning single latent space convolutional filters to the tested semantic concepts.
When applying \gls{cavlrp}, it is apparent that in contrast to \gls{crp}, our approach is more generally applicable, as we do not restrict the used concepts to the encoding of single channels or neurons of the model. 
Therefore, every possible concept that can be encoded with a \gls{cav} can be directly used, additionally opening the box of bias exploration. 
Mined concepts can generally be used by transferring them into linear \gls{cav} vectors, while \gls{crp} can be seen as the special case, where the \gls{cav} is encoded as the unit vector targeting the chosen concept channel only.
As the \gls{crp} concept definition is not bidirectional w.r.t completeness, their concepts are linked to a single channel or neuron inside the model but do not necessarily encapsulate the complete information regarding the semantic (global) concept.
Using \gls{cavlrp} with a \gls{cav} encoding rather focuses on the complete information, which is discriminative w.r.t the appearance of the concept. 

\section{Criteria for the Qualitative Evaluation}

For qualitatively comparing the \gls{cavlrp} explanations e.g. with regard to the concept encoding, we inspected a minimum of 10 different samples per concept and a minimum of 10 different semantic concepts. The comparison between methods is then based on multiple criteria like the clarity of the concept in the explanation, the distribution of attribution over the whole image, the sparseness of the attribution, and the attribution of related features, e.g. attribution on the whole object, when explaining a concept as a single part of the object.

\section{Limitations}

Limitations of our approach are predominantly linked to the quality of concept encodings, which the approach heavily relies on.
While the semantic concept needs to be represented in the learned parameters of the model, a fitting linear encoding of the concept in the correct layer needs to be found.
While \gls{cav} vectors assume a global encoding of the semantic concepts, our research shows that the \gls{cav} only partly encodes the semantic concept.
Specifically, a representation of the concept with limitations in size, orientation, and context is encoded best by the \gls{cav}.
Thus, a medium-sized wheel of a side-facing bicycle might be best reflected in the concept encoding, while the small-sized wheels with arbitrary orientation on a plane or office chair might not be included in the encoding, resulting in bad explanation maps for the \gls{cavlrp} attributions.
\newpage
\section{Additional Visualizations}
\begin{figure*}[h]
\centerline{\includegraphics[width=\linewidth]{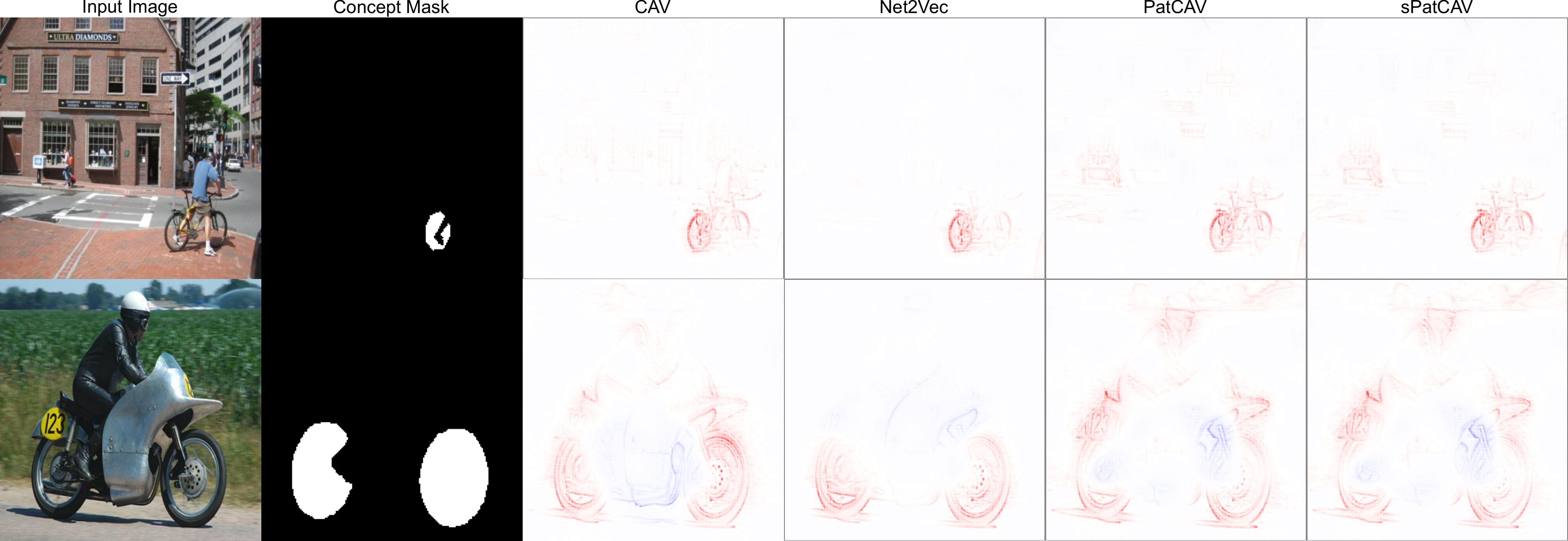}}
\caption{Visual comparison of the \gls{cavlrp} explanations for different global concept encodings in layer \texttt{feat.21} of the SSD model for concept \enquote{wheel}. }
\label{fig:examples}
\end{figure*}

\begin{figure*}[h]
\centerline{\includegraphics[width=\linewidth]{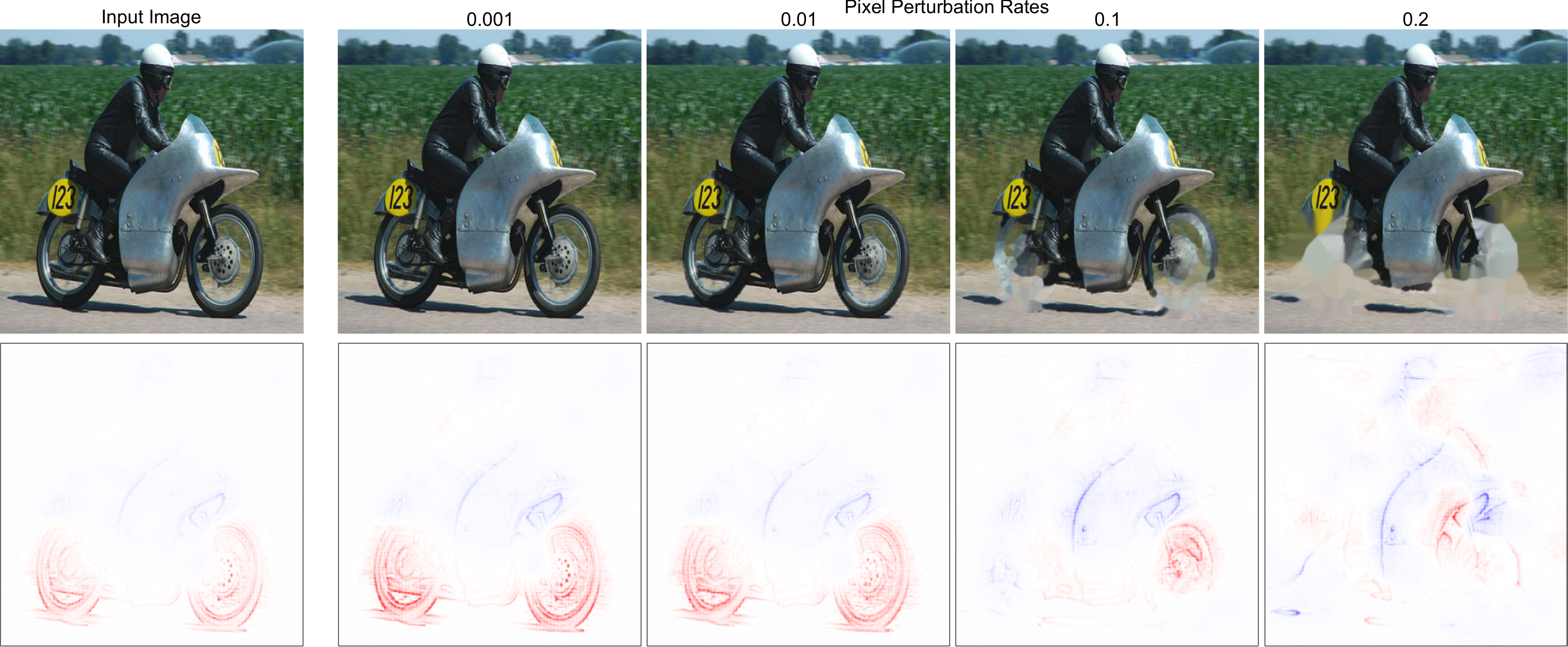}}
\caption{Example of the concept-based input perturbation testing on a single sample using \gls{cavlrp} explanations for concept \enquote{wheel} encoded with \gls{net2vec} in \texttt{feat.21} of the SSD model. }
\label{fig:pixel_pert_example}
\end{figure*}

\begin{figure*}[h]
\centerline{\includegraphics[width=0.7\linewidth]{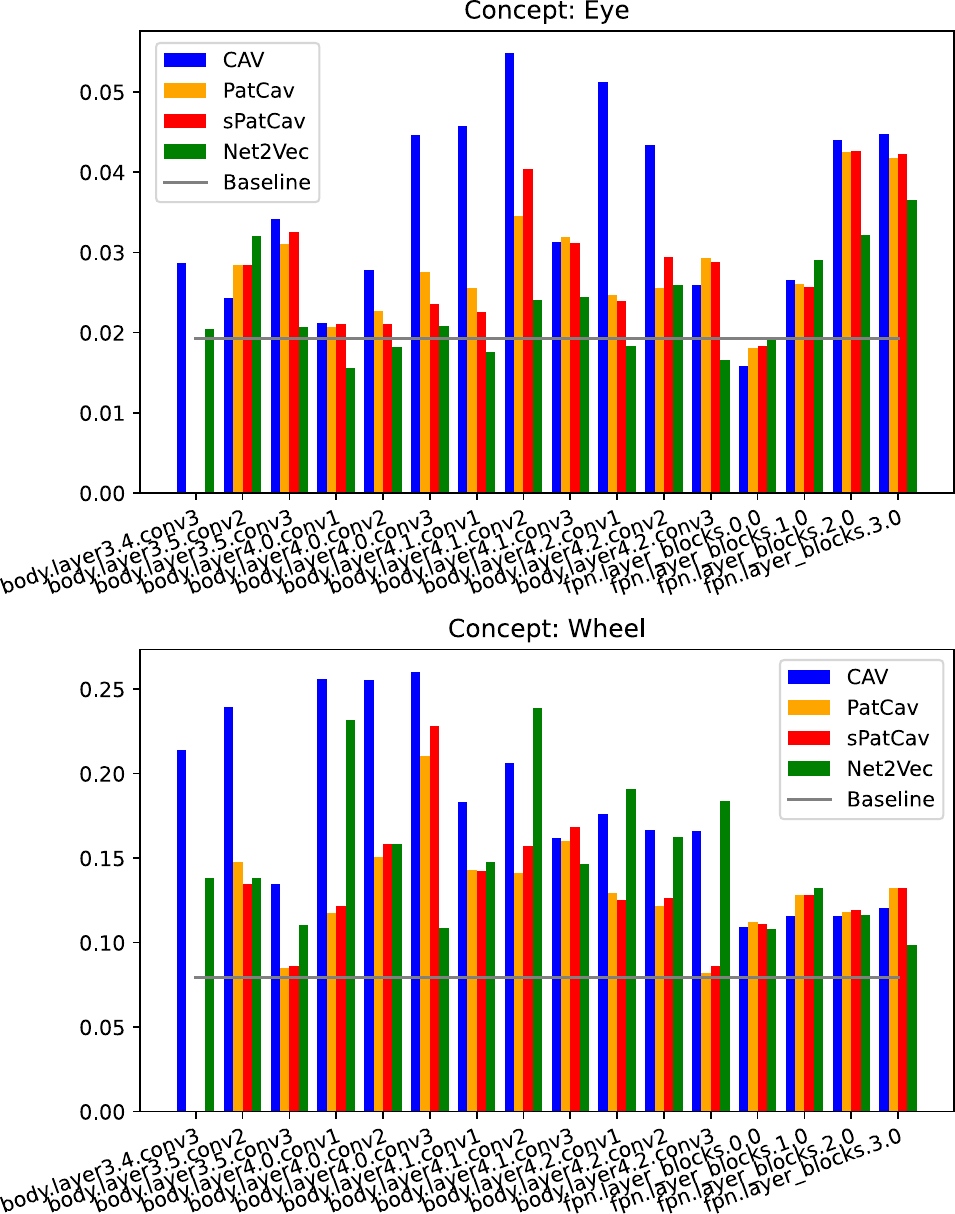}}
\caption{Quantitative attribution localization results for the FasterRCNN model on concepts \enquote{eye} and \enquote{wheel}. }
\label{fig:fcnn_attr_loc}
\end{figure*}

\end{document}